\documentclass{article}
\usepackage{times}
\usepackage{latexsym}
\usepackage{times}  
\usepackage{url}  
\usepackage{graphicx}  
\usepackage{epsfig}
\usepackage{multirow}
\usepackage{booktabs}
\usepackage{amssymb,amsmath}
\usepackage{epsfig}
\usepackage{algorithm,algorithmic}
\usepackage{xcolor,colortbl}
\usepackage{booktabs}
\usepackage{xcolor,colortbl}
\usepackage{makecell}
\usepackage{subcaption}
\usepackage{lipsum}
\usepackage{balance}
\usepackage{authblk}
\newcommand{\eqdef}{\overset{def}{=}}

\title{Unsupervised Dual-Cascade Learning with Pseudo-Feedback Distillation for Query-based Extractive Summarization}

\begin{document}

\author[1]{Haggai Roitman\thanks{Contact author: haggai@il.ibm.com}, Guy Feigenblat, David Konopnicki, Doron Cohen, Odellia Boni}
\affil[1]{IBM Research, Haifa, Israel, 31095}

\date{}

\maketitle
\begin{abstract}
 We propose \textit{Dual-CES} -- a novel unsupervised, query-focused, multi-document extractive summarizer. \textit{Dual-CES} is designed to better handle the tradeoff between saliency and focus in summarization. To this end, \textit{Dual-CES} employs a two-step dual-cascade optimization approach with saliency-based pseudo-feedback distillation. Overall, \textit{Dual-CES} significantly outperforms all other state-of-the-art unsupervised alternatives. 
\textit{Dual-CES} is even shown to be able to outperform strong supervised summarizers.
\end{abstract}

\section{Introduction}\label{sec:intro}
The vast amounts of textual data end users need to consume motivates the need for automatic summarization~\cite{Gambhir:2017}. An automatic summarizer gets as an input one or more documents and possibly also a limit on summary length (e.g., maximum number of words). The summarizer then needs to produce
a textual summary that captures the most  \textit{salient} (general and informative) content parts within input documents.
Moreover, the summarizer may also be required to satisfy a specific user information need, expressed by one or more queries. 
 Therefore, the summarizer will need to produce a \textit{focused} summary which includes the most relevant information to that need. 

\subsection{Motivation}
While both saliency and focus goals should be considered within a query-focused summarization setting, these goals may be actually \textit{conflicting} with each other~\cite{Carbonell:1998:UMD}.
Higher saliency usually comes at the expense of lower focus and vice-versa.
Moreover, such a \textit{tradeoff} may directly depend on summary length.

\begin{figure}[t!]
  \centering
  \includegraphics[width=2.5in]{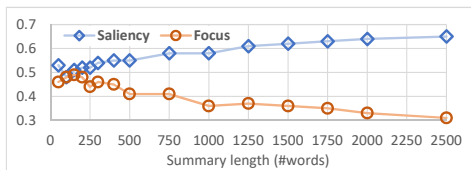}
  \caption{Illustration of the tradeoff between summary saliency and focus goals using varying summary length upper bounds (DUC 2007 dataset).}\label{fig:motivation}
\end{figure}

To illustrate the effect of summary length on this tradeoff, using the DUC 2007 dataset, Figure~\ref{fig:motivation} reports the summarization quality which was obtained by the \textit{Cross Entropy Summarizer} (\textit{CES}) -- a state of the art unsupervised query-focused multi-document extractive summarizer~\cite{summit2017}. 
Saliency was measured according to cosine similarity between the summary's bigram representation and that of the input documents. Focus was further measured relatively to how much the summary's induced unigram model is ``concentrated" around query-related words.

As we can observe in Figure~\ref{fig:motivation}, with the relaxation of the summary length limit, where a more lengthy summary is being allowed, saliency increases at the expense of focus. 
Laying towards more saliency would result in a better coverage of general and more informative content. 
Yet, this would result in the inclusion of less relevant content to the specific information need in mind.

\subsection{Towards a better tradeoff handling}
Aiming at better handling the saliency versus focus tradeoff, in this work, we propose \textit{Dual-CES} -- an extended \textit{CES} summarizer~\cite{summit2017}. 
Similar to \textit{CES}, \textit{Dual-CES} is an \textit{unsupervised} query-focused, multi-document, \textit{extractive} summarizer. To this end, like \textit{CES}, \textit{Dual-CES} utilizes the \textit{Cross Entropy method}~\cite{rubinstein2004} for selecting a subset of sentences extracted from input documents, whose combination is predicted to produce a good summary.

Yet, differently from \textit{CES}, \textit{Dual-CES} does not attempt to address both saliency and focus goals in a single optimization step.
Instead, \textit{Dual-CES} implements a novel two-step dual-cascade optimization approach, which utilizes two sequential \textit{CES}-like invocations. Using such an approach, \textit{Dual-CES} tries to handle the tradeoff by gradually shifting from generating a long summary that is more salient in the first step to generating a short summary that is more focused in the second step. Moreover, \textit{Dual-CES} utilizes the long summary that was generated in the first step for saliency-based \textit{pseudo-feedback distillation}, which allows to generate a final focused summary with better saliency.
\textit{Dual-CES} provides a \textbf{fully unsupervised} end-to-end \textbf{query-focused} multi-document \textbf{extractive} summarization solution.

Using an evaluation with the DUC 2005, 2006 and 2007 benchmarks, we show that, \textit{Dual-CES} generates a focused (and shorter) summary which has much higher saliency (and hence a better tradeoff handling).
Overall, \textit{Dual-CES} provides a significantly better summarization quality compared to other alternative unsupervised summarizers;  
and in many cases, it even outperforms that of state-of-the art supervised summarizers.

\section{Related Work}\label{sec:related}

In this work we employ an unsupervised learning approach for the task of query-based multi-document extractive summarization. 
Many previous works have employed various unsupervised and/or supervised learning methods for the same task. 
Some learning systems rank sentences based on their surface and/or graph level features~\cite{Celikyilmaz:2010,Lin:2011,ouyang2011applying}. Others have used various sparse coding techniques for selecting a subset of sentences that minimizes a given \emph{documents reconstruction error}~\cite{li2017cascaded,IJCAI2015_Spot,COLING2016,He:2012:DSB:2900728.2900817,Li:2015:RMS:2832415.2832426} or used a variational auto-encoder for sentence representation~\cite{li2017salience}. 

Attention models incorporated within deep-learning summarization architectures have further been suggested for improving sentence ranking and selection~\cite{Cao2016AttSumJL,li2017cascaded,Ren:2017LCS}. Such models try to simulate a human attentive reading behaviour. This allows to better account for context-sensitive features during summarization. Compared to these works, we do not try to attend for sentence ranking or selection. Alternatively, we
distill informative hints from summarized documents, aiming to improve the saliency of produced focused summaries. 

Finally, reinforcement learning methods have been recently considered~\cite{chenchen2018,summit2017,Narayan2018,paulus2018deep}. Among such methods, the \textit{CES} summarizer~\cite{summit2017} is the only one which is  both query-sensitive and unsupervised. Similar to \textit{CES}, we also utilize the Cross Entropy (CE) method~\cite{rubinstein2004}, a \textit{global policy search} optimization framework,  for solving the sentence subset selection problem. Yet, differently from \textit{CES}, we utilize the CE method twice, each time with a slightly-different summarization goal in mind (i.e., first saliency and then focus). Moreover, we utilize the distilled saliency-based pseudo-feedback to improve the summarization policy search between such switched (dual) goals. To the best of our knowledge, this on its own, serves as a novel aspect of our work.

\section{Background}\label{sec:background}
Here we provide background details on our summarization task and the \emph{Cross Entropy method} which we use for implementing \textit{Dual-CES}.

\subsection{Summarization task}
We address the query-focused, multi-document summarization task.
Formally, let $q$ denote some user information need for documents summarization, which may be expressed by one or more queries. Let $D$ denote a set of one or more matching documents to be summarized and $L_{max}$ be the maximum allowed summary length (in words).

We implement an extractive summarization approach. 
Our goal is to produce a length-limited summary $S$ by extracting salient content parts in $D$ which are further relevant (focused) to $q$.

Following~\cite{summit2017}, we now cast the summarization task as a \emph{sentence subset selection} problem. To this end, we produce summary $S$ (with maximum length $L_{max}$) by choosing a subset of sentences $s\in{D}$ which maximizes a given quality target $Q(S|q,D)$. 

\subsection{Unsupervised summarization}
\textit{Dual-CES} is an \textit{unsupervised} summarizer. Similar to \textit{CES}, it utilizes the \textit{Cross Entropy method}~\cite{rubinstein2004} for selecting the most ``promising" subset of sentences in $D$. Since we assume an unsupervised setting, no actual reference summaries are available for training nor can we directly optimize an actual quality target $Q(S|q,D)$. Instead, following~\cite{summit2017}, $Q(S|q,D)$ is ``surrogated" by several summary quality prediction measures $\hat{Q}_{i}(S|q,D)$ $(i=1,2,\ldots,m)$. Each ``predictor" $\hat{Q}_{i}(S|q,D)$ is designed to estimate the level of saliency or focus of a given candidate summary $S$ and is presumed to correlate (up to some extent) with actual summarization quality, e.g., ROUGE~\cite{lin2004rouge}. For simplicity, similar to \textit{CES}, various predictions are assumed to be independent and are combined into a single optimization objective by taking their product, i.e.: $\hat{Q}(S|q,D)\eqdef\prod_{i=1}^{m}\hat{Q}_{i}(S|q,D)$. 

\subsection{Using the Cross Entropy method}
The CE-method provides a generic Monte-Carlo optimization framework for solving hard combinatorial problems~\cite{rubinstein2004}. Previously, it was utilized for solving the sentence subset selection problem~\cite{summit2017}.

To this end, the CE-method gets as an input $\hat{Q}(\cdot|q,D)$, a constraint on maximum summary length $L$ and an optional pseudo-reference summary $S_{L}$, whose usage will be explained later on. Let $\texttt{CEM}(\hat{Q}(\cdot|q,D),L,S_{L})$ denote a single invocation of the CE-method. The result of such an invocation is a single length-feasible summary $S^{*}$ which contains a subset of sentences selected from $D$ which maximizes $\hat{Q}(\cdot|q,D)$. For example, \textit{CES} is implemented by invoking $\texttt{CEM}(\hat{Q}_{CES}(\cdot|q,D),L_{max},\emptyset)$.

We next briefly explain how the CE-method solves this problem.
For a given sentence $s\in{D}$, let $\varphi(s)$ denote the likelihood that it should be included in summary $S$. Starting with a selection policy with the highest entropy (i.e.: $\varphi_{0}(s)=0.5$), the CE-Method learns a selection policy $\varphi^{*}(\cdot)$ that maximizes $\hat{Q}(\cdot|q,D)$.
To this end, $\varphi^{*}(\cdot)$ is incrementally learned using an importance sampling approach~\cite{rubinstein2004}.
At each iteration $t=1,2,\ldots$, a sample of $N$ sentence-subsets $S_j$ is generated according to the selection policy $\varphi_{t-1}(\cdot)$ which was learned in the previous iteration $t-1$. The likelihood of picking a sentence $s\in{D}$ at iteration $t$ is estimated (via cross-entropy minimization) as follows: 

 \begin{equation}\label{CE update}
   \varphi_{t}(s)\eqdef\frac{\sum_{j=1}^{N}\delta_{\left[\hat{Q}\left(S_j|q,D\right)\geq\gamma_t\right]}\delta_{\left[s\in{S_j}\right]}}{\sum_{j=1}^{N}\delta_{\left[\hat{Q}\left(S_j|q,D\right)\geq\gamma_t\right]}}.
 \end{equation}

 Here, $\delta_{[\cdot]}$ denotes the \emph{Kronecker-delta} (indicator) function and $\gamma_t$ denotes the $(1-\rho)$-quantile ($\rho\in(0,1)$) of the sample performances $\hat{Q}(S_j|q,D)$ $(j=1,2,\ldots,N)$. Therefore, the likelihood of picking a sentence $s\in D$ will increase when it is being included in more (subset) samples whose performance is above the current minimum required quality target value $\gamma_t$. We further smooth $\varphi_{t}(\cdot)$ as follows: ${\varphi_{t}(\cdot)}'=\alpha \varphi_{t-1}(\cdot)+(1-\alpha)\varphi_{t}(\cdot)$; with $\alpha\in[0,1]$~\cite{rubinstein2004}.

 Upon its termination, the CE-method is expected to converge to the global optimal selection policy $\varphi^{*}(\cdot)$~\cite{rubinstein2004}. We then produce a single summary $S^{*}\sim\varphi^{*}(\cdot)$.
 To enforce that only feasible summaries will be produced, following~\cite{summit2017}, we set $\hat{Q}(S_j|q,D)=-\infty$ whenever a sampled summary $S_j$ length exceeds the $L$ word limit. 

 \section{The Dual-CES summarizer}\label{sec:solution}
\begin{figure}[t!]
  \centering
  \includegraphics[width=3.0in]{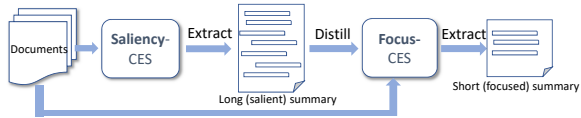}
  \caption{Dual-CES implementation flow}\label{fig:flow}
\end{figure}

Differently from \textit{CES}, \textit{Dual-CES} does not attempt to maximize both saliency and focus goals in a single optimization step.
Instead, \textit{Dual-CES} implements a novel two-step dual-cascade optimization approach (see Figure~\ref{fig:flow}), which utilizes two \textit{CES}-like invocations. Both invocations consider the \textbf{same} sentences powerset solution space. Yet, each such invocation utilizes a bit different set of summary quality predictors $\hat{Q}_{i}(S|q,D)$, depending on whether the summarizer's goal should lay towards higher summary saliency or focus.

In the first step, \textit{Dual-CES} relaxes the 
summary length constraint, aiming at producing a longer and more salient summary. This summary is then treated as a \textit{pseudo-effective reference summary} from which \textit{saliency-based pseudo-feedback} is \textit{distilled}. Such pseudo-feedback is then utilized in the second step of the cascade for setting an additional auxiliary saliency-driven goal. Yet, at the second step, similar to \textit{CES}, the primary goal is actually to produce a focused summary (with maximum length limit $L_{max}$). Overall, \textit{Dual-CES} is simply implemented as follows:
\[
\texttt{CEM}(\hat{Q}_{Foc}(\cdot|q,D),L_{max},\texttt{CEM}(\hat{Q}_{Sal}(\cdot|q,D),\bar{L},\emptyset)).
\]

Here, $\hat{Q}_{Sal}(\cdot|q,D)$ and $\hat{Q}_{Foc}(\cdot|q,D)$ denote the saliency and focus summary quality objectives which are optimized during the cascade, respectively. Both $\hat{Q}_{Sal}(\cdot|q,D)$ and $\hat{Q}_{Foc}(\cdot|q,D)$ are implemented as a product of several basic predictors. $\bar{L}\geq L_{max}$ denotes the relaxed summary length hyperparameter.
We next elaborate the implementation details of \textit{Dual-CES}'s dual optimization steps.

\subsection{Step 1: Saliency-oriented summarization}
The purpose of the first step is to produce a single longer summary (with length $\bar{L}\geq{L_{max}}$) which will be used as a pseudo-reference for saliency-based feedback distillation. 
As illustrated in Figure~\ref{fig:motivation}, with a longer summary length -- a more salient summary may be produced.

This step is simply implemented by invoking the CE-method with $\texttt{CEM}(\hat{Q}_{Sal}(\cdot|q,D),\bar{L},\emptyset)$. The target measure $\hat{Q}_{Sal}(\cdot|q,D)$ guides the optimization towards the production of a summary with the highest possible saliency. Similar to \textit{CES}, $\hat{Q}_{Sal}(\cdot|q,D)$ is calculated as the product of several summary quality predictors. Overall, we use five different predictors, four of which were previously used in \textit{CES}~\cite{summit2017}. The additional predictor that we introduce is designed to ``drive" the optimization even further towards higher saliency. Next, we shortly describe each predictor. The symbol $\dag$ marks whether it was originally employed in \textit{CES}~\cite{summit2017}.

\subsubsection{Predictor 1: coverage$^\dag$}
This predictor estimates to what extent (candidate) summary $S$ (generally) covers the document set $D$. Here, we represent both $S$ and $D$ as term-frequency vectors, considering only bigrams, which commonly represent more important content units~\cite{summit2017}. For a given text $x$, let $cos(S,x)\eqdef\frac{\vec{S}\cdot\vec{x}}{\|\vec{S}\|\|\vec{x}\|}$. The coverage predictor is then defined by $\hat{Q}_{cov}(S|q,D)\eqdef cos(S,D)$.

\subsubsection{Predictor 2: position-bias$^\dag$}
This predictor biases sentence selection towards sentences that appear earlier in their containing documents.
It is calculated as $\hat{Q}_{pos}(S|q,D)\eqdef\sqrt[|S|]{\prod\limits_{s\in{S}}\left(1+\frac{1}{\log(b+pos(s))}\right)}$, where $pos(s)$ is the relative start position (in characters) of sentence $s$ in its containing document and $b$ is a \textit{position-bias} hyperparameter (fixed to $b=2$, following~\cite{summit2017}).

\subsubsection{Predictor 3: summary length$^\dag$}
This predictor biases towards selection of summaries that are closer to the maximum permitted length. Such summaries contain fewer and longer sentences, and therefore, tend to be more informative. Let $len(x)$ denote the length of text $x$ (in number of words). Here, $x$ may either be a single sentence $s\in{D}$ or a whole summary $S$. This predictor is then calculated as $\hat{Q}_{len}(S|q,D)\eqdef\frac{1}{|S|}len(S)$, where $len(S)=\sum_{s\in{S}}len(s)$.

\subsubsection{Predictor 4: asymmetric coverage}
To target even higher saliency, we suggest a fourth predictor, inspired by the \textit{risk minimization framework}~\cite{Zhai:2006:RMF}.  To this end, we measure the \textit{Kullback-Leibler} (KL) ``similarity" between the two (unsmoothed) unigram language models induced from the centroid representation\footnote{Such centroid representation is simply given by concatenating the text of sentences in $S$ or documents in $D$.}  of $S$ ($\hat{\theta}_{S}$) and $D$ ($\hat{\theta}_{D}$), formally: 

$\hat{Q}_{KL}(S|q,D)\eqdef\exp\left({-\sum_{w}p(w|\hat{\theta}_{S})\log\frac{p(w|\hat{\theta}_{S})}{p(w|\hat{\theta}_{D})}}\right)$.

\subsubsection{Predictor 5: focus-drift$^\dag$}\label{sec:query focus}
While producing a longer summary may result in higher saliency, as was further illustrated in Figure~\ref{fig:motivation}, such a summary may be less focused. Hence, to avoid such \textit{focus-drift}, while we opt to optimize for higher saliency at this step, the target information need $q$ should be still considered.
To this end, we add a 
predictor: $\hat{Q}_{qf}(S|q,D)\eqdef\sum\limits_{w\in{q}}p(w|\hat{\theta}_{S})$, which acts as a ``query-anchor" and measures to what extent summary $S$'s unigram model is devoted to the information need $q$.

\subsection{Step 2: Focus-oriented summarization}
The input to the second step of the cascade consists of the \textbf{same} set of documents $D$, summary length constraint $L_{max}$ and the pseudo-reference summary $S_{\bar{L}}$ that was generated in the previous step.
This step is simply implemented by invoking the CE-method with $\texttt{CEM}(\hat{Q}_{Foc}(\cdot|q,D),L_{max},S_{\bar{L}})$. Here, the target measure $\hat{Q}_{Foc}(\cdot|q,D)$ guides the optimization towards the production of a focused summary, while still keeping high saliency as much as possible. To achieve that, we use an additional focus-driven predictor which bias summary production towards higher focus. Moreover, using the pseudo-reference summary $S_{\bar{L}}$ we introduce an additional auxiliary saliency-based predictor, whose goal is to enhance the saliency of produced focused summary. 
Overall, $\hat{Q}_{Foc}(\cdot|q,D)$ is calculated as the product of the previous five summary quality predictors (Predictors 1$-$5) 
and the two additional predictors, whose details are described next.

\subsubsection{Predictor 6: query-relevancy$^\dag$}
This predictor estimates the relevancy of summary $S$ to $q$. For that, we use two similarity measures. The first, following~\cite{summit2017},  measures the \emph{Bhattacharyya} similarity (coefficient) between the two (unsmoothed) unigram language models of $q$ and $S$, i.e.: $\hat{Q}_{sim1}(S|q,D)\eqdef\sum_{w\in{q}}\sqrt{p(w|\hat{\theta}_{q})p(w|\hat{\theta}_{S})}$.
The second measures the cosine similarity between $q$ and $S$ unigram term-frequency representations, i.e.: $\hat{Q}_{sim2}(S|q,D)\eqdef cos(S,q)$. The two similarity measures are then combined into a single measure using their geometric mean, i.e.: $\hat{Q}_{sim}(S|q,D)\eqdef \sqrt{\hat{Q}_{sim1}(S|q,D)\cdot \hat{Q}_{sim2}(S|q,D)}$.

\subsubsection{Predictor 7: reference summary (distillation) coverage}
We further make use of the pseudo-reference summary $S_{\bar{L}}$, which was produced in the first step, and introduce an additional auxiliary saliency-based predictor. This predictor utilizes pseudo-feedback that is distilled from unique unigram words in $S_{\bar{L}}$.  It is calculated as:
$\hat{Q}_{cov'}(S|q,D)\eqdef\sum_{w\in{S_{\bar{L}}}}\delta_{[w\in{S}]}$. Following~\cite{Lavrenko:2001,Zhai:2006:RMF}, we only consider the top-100 most frequent unigrams in $S_{\bar{L}}$.

Intuitively speaking, $S_{\bar{L}}$ usually will be longer (in words) than any candidate summary $S$ that may be chosen in the second step; hence, $S_{\bar{L}}$ is expected to be more salient than $S$. Therefore, such a predictor is expected to ``drive" the optimization to prefer those candidate summaries $S$ that include as many salient words from $S_{\bar{L}}$, acting as if they were by themselves longer (and more salient) summaries (than those candidates that include less salient words from $S_{\bar{L}}$).

\subsubsection{Adaptive hyperparameter $b$ adjustment}
Apart from salient words in $S_{\bar{L}}$ that are used as feedback, we note that, sentences in $S_{\bar{L}}$ may also provide additional ``hints" about other properties of informative sentences in $D$, which may potentially be selected to improve saliency. One such property is the relative start-positions of sentences in $S_{\bar{L}}$. To this end, 
we now assign $b=\frac{1}{|S_{\bar{L}}|}\sum_{s\in S_{\bar{L}}}pos(s)$ (i.e., the average start-position of feedback sentences in $S_{\bar{L}}$) as the value of the position-bias hyperparameter within $\hat{Q}_{pos}(S|q,D)$ (Predictor 2).

\subsection{An extension: Length-adaptive Dual-CES}\label{sec:apative length}
We conclude this section with a suggestion of an extension to \textit{Dual-CES} that adaptively adjusts the value of hyperparameter $\bar{L}$. %
To this end, we introduce a new learning parameter $L_{t}$ which defines the maximum length limit for summary production (sampling) that is allowed at iteration $t$ of the CE-method.
We now assume that summary lengths have a \textit{Poisson}$(L_{t})$ distribution of word occurrences with mean $L_{t}$. Using importance sampling, this parameter is estimated at iteration $t$ as follows:
 \begin{equation}\label{adaptive L}
   L_{t}\eqdef\frac{\sum_{j=1}^{N}len(S_j)\cdot\delta_{\left[\hat{Q}\left(S_j|q,D\right)\geq\gamma_t\right]}}{\sum_{j=1}^{N}\delta_{\left[\hat{Q}\left(S_j|q,D\right)\geq\gamma_t\right]}}.
 \end{equation}

 Similar to $\varphi(\cdot)$, we further smooth $L_{t}$ as follows: ${L_{t}}'\eqdef\alpha L_{t-1}+(1-\alpha)L_t$. Here, $\alpha\in[0,1]$ is the same smoothing hyperparameter which was used to smooth $\varphi(\cdot)$ and $L_{t=0}\eqdef{\bar{L}}$.

\section{Evaluation}\label{sec:eval}

\subsection{Datasets}
Our evaluation is based on the \emph{Document Understanding Conferences} (DUC) 2005, 2006 and 2007 benchmarks\footnote{\url{http://www-nlpir.nist.gov/projects/duc/data.html}}. These benchmarks are commonly used for evaluating the query-based multi-document summarization task by all of our related works.
Given a topic statement, which is expressed by one or more questions, and a set of English documents, the main task is to produce a 250-word (i.e., $L_{max}=250$) topic-focused summary~\cite{dang2005overview}. The number of topics per benchmark are $50$, $50$ and $45$ in the DUC 2005, 2006 and 2007 benchmarks, respectively. The number of documents to be summarized per topic is $32$, $25$ and $25$ in the DUC 2005, 2006 and 2007 benchmarks, respectively. Each document was pre-segmented (by NIST) into sentences. Following~\cite{summit2017}, we use Lucene's English analysis\footnote{\url{https://lucene.apache.org/}} for processing the text of topics and documents.

\subsubsection{Dual-CES implementation}
We evaluated both \textit{Dual-CES} and its adaptive-length variant (hereinafter denoted \textit{Dual-CES-A}). To this end, on the first saliency-driven step, for \textit{Dual-CES}, we fixed the (strict) upper bound limit on summary length to $\bar{L}=1500$.  \textit{Dual-CES-A}, on the other hand, adaptively adjusts such length limit and was initialized with $L_{t=0}=3000$. Both variants were further set with a summary limit $L_{max}=250$ for their second focus-driven step.

 We implemented both \textit{Dual-CES} and \textit{Dual-CES-A} in Java (JRE8).
Further following~\cite{summit2017}, to reduce CE-method's runtime, we applied a preliminary step of sentence pruning, where only the top-150 sentences $s\in{D}$ with the highest (unigram) \emph{Bhattacharyya} similarity to the topic's queries were considered for summarization. Similar to~\cite{summit2017}, 
the CE-method hyperparameters were fixed as follows: $N=10,000$, $\rho=0.01$ and $\alpha=0.7$.

Finally, to handle DUC's complex information needs, we closely followed~\cite{summit2017}, as follows. First, for each summarized topic, we calculated the query-focused predictions (i.e., $\hat{Q}_{qf}(\cdot|q,D)$ and $\hat{Q}_{sim}(\cdot|q,D)$) per each one of its questions. To this end, each question was represented as a sub-query by concatenating the main topic's text to the question's text. Each sub-query was further expanded with top-$100$ (unigram) Wikipedia related-words~\cite{Xu:2009}. We then obtained the topic query-sensitive predictions by summing up its various sub-queries' predictions.

\subsubsection{Evaluation measures}
The three DUC benchmarks include four reference (ground-truth) human-written summaries per each topic~\cite{dang2005overview}. We measured summarization quality using the ROUGE measure~\cite{lin2004rouge}, which is the official one for this task~\cite{dang2005overview}. To this end, we used the ROUGE $1.5.5$ toolkit with its standard parameters setting\footnote{ROUGE-1.5.5.pl -a -c 95 -m -n 2 -2 4 -u -p 0.5 -l 250}. We report both Recall and F-Measure of ROUGE-1, ROUGE-2 and ROUGE-SU4.
ROUGE-1 and ROUGE-2 measure the overlap in unigrams and bigrams between the produced and the reference summaries, respectively. 
ROUGE-SU4 measures the overlap in skip-grams separated by up to four words. 

Finally, since \textit{Dual-CES} essentially depends on the CE-method which has a stochastic nature, its quality may depend on the specific seed that was used for random sampling. Hence, following~\cite{summit2017}, to reduce sensitivity to random seed selection, per each summarization task (i.e., topic and documents pair), we run each \textit{Dual-CES} variant 30 times (each time with a different random seed) and recorded its mean performance (and $95\%$ confidence interval).

\subsection{Baselines}
We compare the summary quality of \textit{Dual-CES} to the results that were previously reported for several competitive summarization baselines. These baselines include both supervised and unsupervised methods and apply various strategies for handling the saliency versus focus tradeoff. To distinguish between both types of works, we mark supervised method names with a superscript $\S$.

The first line of baselines utilize various surface and graph level features, namely: \textit{BI-PLSA}~\cite{Shen:2011:ICM:2900423.2900569},  \textit{CTSUM} ~\cite{Wan:2014:CEM:2600428.2609559}, \textit{HierSum}~\cite{Haghighi:2009}, \textit{HybHSum}$^\S$~\cite{Celikyilmaz:2010}, \textit{MultiMR}~\cite{Wan:2009:GML:1661445.1661700}, \textit{QODE}~\cite{Zhong:2015:QUM:2827896.2828239} and \textit{SubMod-F}$^\S$~\cite{Lin:2011}. The second line of baselines apply various sparse-coding or auto-encoding techniques, namely: \textit{DocRebuild}~\cite{COLING2016}, \textit{RA-MDS}~\cite{Li:2015:RMS:2832415.2832426}, \textit{SpOpt}~\cite{IJCAI2015_Spot},  and \textit{VAEs-A}~\cite{li2017salience}. The third line of baselines incorporate various attention models, namely: \textit{AttSum}$^\S$~\cite{Cao2016AttSumJL}, \textit{C-Attention}~\cite{li2017cascaded} and \textit{CRSum+SF}$^\S$~\cite{Ren:2017LCS}. We further note that, some baselines, like \textit{DocRebuild}, \textit{SpOpt} and \textit{C-Attention}, use hand-crafted rules for sentence compression.

Finally, we directly compare with two \textit{CES} variants, which serve as direct alternatives to \textit{Dual-CES}. The first one, is the original \textit{CES} summarizer, whose results are reported in~\cite{summit2017}. The second one, denoted hereinafter \textit{CES}$^{+}$, utilizes Predictors 1$-$6, which are combined within a single optimized objective (by taking their product). This variant, therefore, allows to directly evaluate the contribution of our proposed dual-cascade learning approach which is employed by the two \textit{Dual-CES} variants.

\subsection{Results}
The main results of our evaluation are reported in Table~\ref{tab:ROUGE-F} (ROUGE-X F-Measure) and Table~\ref{tab:ROUGE-R} (ROUGE-X Recall). The numbers reported for the various baselines are the \textbf{best} numbers reported in their respective works. Unfortunately,  not all baselines fully reported their results for all benchmarks and measures. Whenever a report on a measure is missing, we further use the symbol '-'. 

\begin{table}[tbh!]
\small
\centering
\begin{tabular}{@{}lllll@{}}
\toprule
                                                           & System      & R-1                                                                & R-2                                                               & R-SU4                                                              \\ \midrule
\multirow{4}{*}{\begin{tabular}[c]{@{}l@{}}DUC\\ 2005\end{tabular}} & MultiMR     & 36.90                                                              & 6.83                                                              & -                                                                  \\
                                                                    & CES  & \begin{tabular}[c]{@{}c@{}}37.76\tiny{($\pm.$03)}\end{tabular}           & \begin{tabular}[c]{@{}c@{}}7.45\tiny{($\pm.$03)}\end{tabular}           & \begin{tabular}[c]{@{}c@{}}13.02\tiny{($\pm.$02)}\end{tabular}           \\
                                                                    & CES$^{+}$  & \begin{tabular}[c]{@{}c@{}}36.94\tiny{($\pm.$01)}\end{tabular} & \begin{tabular}[c]{@{}c@{}}7.21\tiny{($\pm.$04)}\end{tabular} & \begin{tabular}[c]{@{}c@{}}12.82\tiny{($\pm.$04)}\end{tabular} \\
																																		& Dual-CES-A  & \textbf{\begin{tabular}[c]{@{}c@{}}38.13\tiny{($\pm.$07)}\end{tabular}} & \textbf{\begin{tabular}[c]{@{}c@{}}7.58\tiny{($\pm.$04)}\end{tabular}} & \textbf{\begin{tabular}[c]{@{}c@{}}13.24\tiny{($\pm.$04)}\end{tabular}} \\
                                                                    & Dual-CES    & \begin{tabular}[c]{@{}c@{}}38.08\tiny{($\pm.$06)}\end{tabular}          & \begin{tabular}[c]{@{}c@{}}7.54\tiny{($\pm.$03)}\end{tabular}          & \begin{tabular}[c]{@{}c@{}}13.17\tiny{($\pm.$03)}\end{tabular}      \\ \midrule
\multirow{8}{*}{\begin{tabular}[c]{@{}l@{}}DUC\\ 2006\end{tabular}} & RA-MDS      & 39.10                                                               & 8.10                                                               & 13.6
                                                                    \\
                                                                    & MultiMR     & 40.30                                                              & 8.50                                                              & -                                                                 \\
                                                                    & DocRebuild  & 40.86                                                              & 8.48                                                              & 14.45                                                              \\
                                                                    & C-Attention & 39.30                                                               & 8.70                                                               & 14.10                                                               \\
                                                                    & VAEs-A      & 39.60                                                               & 8.90                                                               & 14.30                                                               \\
                                                                    & CES  & \begin{tabular}[c]{@{}c@{}}40.46\tiny{($\pm.$02)}\end{tabular}           & \begin{tabular}[c]{@{}c@{}}9.13\tiny{($\pm.$01)}\end{tabular}           & \begin{tabular}[c]{@{}c@{}}14.71\tiny{($\pm.$01)}\end{tabular} \\
                                                                    & CES$^{+}$  & \begin{tabular}[c]{@{}c@{}}39.93\tiny{($\pm.$08)}\end{tabular}           & \begin{tabular}[c]{@{}c@{}}9.02\tiny{($\pm.$05)}\end{tabular}           & \begin{tabular}[c]{@{}c@{}}14.42\tiny{($\pm.$05)}\end{tabular}           \\
																																		& Dual-CES-A  & \begin{tabular}[c]{@{}c@{}}41.07\tiny{($\pm.$07)}\end{tabular}           & \begin{tabular}[c]{@{}c@{}}9.42\tiny{($\pm.$06)}\end{tabular}           & \begin{tabular}[c]{@{}c@{}}14.89\tiny{($\pm.$05)}\end{tabular}           \\
                                                                    & Dual-CES    & \textbf{\begin{tabular}[c]{@{}c@{}}41.23\tiny{($\pm.$07)}\end{tabular}}  & \textbf{\begin{tabular}[c]{@{}c@{}}9.47\tiny{($\pm.$04)}\end{tabular}}  & \textbf{\begin{tabular}[c]{@{}c@{}}14.97\tiny{($\pm.$03)}\end{tabular}}  \\ \midrule
\multirow{9}{*}{\begin{tabular}[c]{@{}l@{}}DUC\\ 2007\end{tabular}}
                                                                    & RA-MDS      & 40.80                                                               & 9.70                                                               & 15.00                                                                 \\
                                                                    & MultiMR     & 42.04                                                              & 10.30                                                             & -                                                                  \\
                                                                    & DocRebuild  & 42.72                                                              & 10.30                                                             & 15.81                                                              \\
                                                                    & CTSUM       & 42.66                                                              & 10.83                                                             & 16.16                                                              \\
                                                                    & C-Attention & 42.30                                                               & 10.70                                                              & 16.10                                                               \\
                                                                    & VAEs-A      & 42.10                                                              & 11.10                                                              & 16.40                                                               \\
                                                                    & CES  & \begin{tabular}[c]{@{}c@{}}42.84\tiny{($\pm.$01)}\end{tabular}           & \begin{tabular}[c]{@{}c@{}}11.33\tiny{($\pm.$01)}\end{tabular}           & \begin{tabular}[c]{@{}c@{}}16.50\tiny{($\pm.$01)}\end{tabular}           \\
																																		& CES$^{+}$  & \begin{tabular}[c]{@{}c@{}}41.90\tiny{($\pm.$08)}\end{tabular}           & \begin{tabular}[c]{@{}c@{}}11.14\tiny{($\pm.$06)}\end{tabular}           & \begin{tabular}[c]{@{}c@{}}16.17\tiny{($\pm.$05)}\end{tabular}           \\
                                                                    & Dual-CES-A  & \textbf{\begin{tabular}[c]{@{}c@{}}43.25\tiny{($\pm.$06)}\end{tabular}}           & \begin{tabular}[c]{@{}c@{}}11.73\tiny{($\pm.$06)}\end{tabular}          & \begin{tabular}[c]{@{}c@{}}16.80\tiny{($\pm.$04)}\end{tabular}            \\
                                                                    & Dual-CES    & \begin{tabular}[c]{@{}c@{}}43.24\tiny{($\pm.$07)}\end{tabular}  & \textbf{\begin{tabular}[c]{@{}c@{}}11.78\tiny{($\pm.$05)}\end{tabular}} & \textbf{\begin{tabular}[c]{@{}c@{}}16.83\tiny{($\pm.$05)}\end{tabular}}  \\ \bottomrule
\end{tabular}
\caption{Results of ROUGE F-Measure evaluation on DUC 2005, 2006, and 2007 benchmarks.}
\label{tab:ROUGE-F}
\end{table}

\begin{table}[tbh!]
\small
\centering
\begin{tabular}{@{}lllll@{}}
\toprule
                                                              & System     & R-1                                                               & R-2                                                               & R-SU4                                                             \\ \midrule
\multirow{6}{*}{\begin{tabular}[c]{@{}l@{}}DUC \\ 2005\end{tabular}} & SubMod-F$^\S$   & -                                                                 & 8.38                                                              & -                                                                 \\
                                                                     & CRSum+SF$^\S$      & 39.52                                                             & \textbf{8.41}                                                              & -                                                                 \\ \cmidrule(l){2-5}
                                                                     & BI-PLSA    & 36.02                                                             & 6.76                                                              & -                                                                 \\
                                                                     & CES  & \begin{tabular}[c]{@{}c@{}}40.33\tiny{($\pm.$03)}\end{tabular}   & \begin{tabular}[c]{@{}c@{}}7.94\tiny{($\pm.$02)}\end{tabular}  & \begin{tabular}[c]{@{}c@{}}13.89\tiny{($\pm.$02)}\end{tabular}           \\
																																		& CES$^{+}$  & \begin{tabular}[c]{@{}c@{}}39.56\tiny{($\pm.$11)}\end{tabular} & \begin{tabular}[c]{@{}c@{}}7.71\tiny{($\pm.$04)}\end{tabular}  & \begin{tabular}[c]{@{}c@{}}13.73\tiny{($\pm.$05)}\end{tabular}	\\
																																		
                                                                     & Dual-CES-A & \textbf{\begin{tabular}[c]{@{}c@{}}40.85\tiny{($\pm.$07)}\end{tabular}}& \begin{tabular}[c]{@{}c@{}}8.11\tiny{($\pm.$04)}\end{tabular}  & \textbf{\begin{tabular}[c]{@{}c@{}}14.19\tiny{($\pm.$04)}\end{tabular}}\\
                                                                     & Dual-CES   & \begin{tabular}[c]{@{}c@{}}40.82\tiny{($\pm.$06)}\end{tabular}          & \begin{tabular}[c]{@{}c@{}}8.07\tiny{($\pm.$04)}\end{tabular}           & \begin{tabular}[c]{@{}c@{}}14.13\tiny{($\pm.$04)}\end{tabular}          \\ \midrule
\multirow{9}{*}{\begin{tabular}[c]{@{}l@{}}DUC\\ 2006\end{tabular}}
                                                                     & AttSum$^\S$     & 40.90                                                             & 9.40                                                              & -                                                                 \\
                                                                     & SubMod-F$^\S$   & -                                                                 & 9.75                                                              & -                                                                 \\
                                                                     & HybHSum$^\S$    & 43.00                                                             & 9.10                                                              & 15.10                                                             \\
                                                                     & CRSum+SF$^\S$      & 41.70                                                             & 10.03                                                             & -                                                                 \\ \cmidrule(l){2-5}
                                                                     & HierSum    & 40.10                                                             & 8.60                                                              & 14.30                                                             \\
                                                                     & SpOpt      & 39.96                                                             & 8.68                                                              & 14.22                                                             \\
                                                                     & QODE       & 40.15                                                             & 9.28                                                              & 14.79                                                             \\
                                                                     & CES  & \begin{tabular}[c]{@{}c@{}}43.00\tiny{($\pm.$01)}\end{tabular}           & \begin{tabular}[c]{@{}c@{}}9.69\tiny{($\pm.$01)}\end{tabular}           & \begin{tabular}[c]{@{}c@{}}15.63\tiny{($\pm.$01)}\end{tabular}           \\
																																			& CES$^{+}$ & \begin{tabular}[c]{@{}c@{}}42.57\tiny{($\pm.$09)}\end{tabular}     & \begin{tabular}[c]{@{}c@{}}9.61\tiny{($\pm.$06)}\end{tabular}          & \begin{tabular}[c]{@{}c@{}}15.38\tiny{($\pm.$06)}\end{tabular}          		\\
                                                                     & Dual-CES-A & \begin{tabular}[c]{@{}c@{}}43.78\tiny{($\pm.$07)}\end{tabular}          & \begin{tabular}[c]{@{}c@{}}10.04\tiny{($\pm.$06)}\end{tabular}          & \begin{tabular}[c]{@{}c@{}}15.88\tiny{($\pm.$05)}\end{tabular}          \\
                                                                     & Dual-CES   & \textbf{\begin{tabular}[c]{@{}c@{}}43.94\tiny{($\pm.$07)}\end{tabular}} & \textbf{\begin{tabular}[c]{@{}c@{}}10.09\tiny{($\pm.$05)}\end{tabular}} & \textbf{\begin{tabular}[c]{@{}c@{}}15.96\tiny{($\pm.$03)}\end{tabular}} \\ \midrule
\multirow{9}{*}{\begin{tabular}[c]{@{}l@{}}DUC \\ 2007\end{tabular}} & AttSum$^\S$     & 43.92                                                             & 11.55                                                             & -                                                                 \\
                                                                     & SubMod-F$^\S$   & -                                                                 & 12.38                                                             & -                                                                 \\
                                                                     & HybHSum$^\S$    & 45.60                                                             & 11.40                                                              & 17.20                                                             \\
                                                                     & CRSum+SF$^\S$      & 44.60                                                             & 12.48                                                             & -                                                                 \\
                                                                     \cmidrule(l){2-5}
                                                                     & HierSum    & 42.40                                                             & 11.80                                                              & 16.70                                                             \\
                                                                     & SpOpt      & 42.36                                                             & 11.10                                                             & 16.47                                                             \\
                                                                     & QODE       & 42.95                                                             & 11.63                                                             & 16.85                                                             \\
                                                                     & CES  & \begin{tabular}[c]{@{}c@{}}45.43\tiny{($\pm.$01)}\end{tabular}           & \begin{tabular}[c]{@{}c@{}}12.02\tiny{($\pm.$01)}\end{tabular}           & \begin{tabular}[c]{@{}c@{}}17.50\tiny{($\pm.$01)}\end{tabular}           \\
																																		 & CES$^{+}$   & \begin{tabular}[c]{@{}c@{}}44.65\tiny{($\pm.$01)}\end{tabular}           & \begin{tabular}[c]{@{}c@{}}11.85\tiny{($\pm.$01)}\end{tabular}           & \begin{tabular}[c]{@{}c@{}}17.21\tiny{($\pm.$06)}\end{tabular}   				\\
																																		& Dual-CES-A & \begin{tabular}[c]{@{}c@{}}46.01\tiny{($\pm.$07)}\end{tabular}          & \begin{tabular}[c]{@{}c@{}}12.47\tiny{($\pm.$06)}\end{tabular}          & \begin{tabular}[c]{@{}c@{}}17.87\tiny{($\pm.$04)}\end{tabular}          \\
                                                                     & Dual-CES   & \textbf{\begin{tabular}[c]{@{}c@{}}46.02\tiny{($\pm.$08)}\end{tabular}} & \textbf{\begin{tabular}[c]{@{}c@{}}12.53\tiny{($\pm.$06)}\end{tabular}} & \textbf{\begin{tabular}[c]{@{}c@{}}17.91\tiny{($\pm.$05)}\end{tabular}} \\ \cmidrule(l){2-5}
\end{tabular}

\caption{Results of ROUGE Recall evaluation on DUC 2005, 2006, and 2007 benchmarks.} 
\label{tab:ROUGE-R}
\end{table}

\subsubsection{Dual-CES vs. other baselines}
First we note that, among the various baseline methods that we have compared with, \textit{CES} on its own, serves as the strongest baseline to outperform in most cases. 
Overall, \textit{Dual-CES} provides better results compared to any other baseline (and specifically the unsupervised ones). 
Specifically, on F-Measure, \textit{Dual-CES} has achieved between $6\%-14\%$ and $1\%-3\%$ better ROUGE-2 and ROUGE-1, respectively. On recall, \textit{Dual-CES} has achieved between $3\%-9\%$ better ROUGE-1. On ROUGE-2, in the DUC 2006 and 2007 benchmarks, \textit{Dual-CES} was about $1\%-9\%$ better, while it was slightly inferior to \textit{SubMod-F} and \textit{CRSum+SF} in the DUC 2005 benchmark. Yet, \textit{SubMod-F} and \textit{CRSum+SF} are actually supervised, while \textit{Dual-CES} is fully unsupervised. Therefore, overall, \textit{Dual-CES}'s ability to reach (even to outperform in many cases) the quality of strong supervised counterparts actually only emphasizes more its potential.

\subsubsection{Dual-CES vs. CES variants}
\textit{Dual-CES} significantly improves over the two \textit{CES} variants in all benchmarks.
On F-Measure, \textit{Dual-CES} has achieved at least between $4\%-5\%$  and $1\%-2\%$ better ROUGE-2 and ROUGE-1, respectively. On recall, \textit{Dual-CES} has achieved at least between $2\%-4\%$ and $1\%-2\%$  better ROUGE-2 and ROUGE-1, respectively. 
By distilling saliency-based pseudo-feedback between step transitions, \textit{Dual-CES} manages to better utilize the CE-method for selecting a more promising subset of sentences. A case in point is the \textit{CES}$^{+}$ variant which is even inferior to \textit{CES}.  A simple combination of all predictors (except Predictor 7 which is unique to \textit{Dual-CES} since it requires a pseudo-reference summary) does not directly translates to a better tradeoff handling. This, therefore, serves as a strong empirical evidence of the importance of the dual-cascade optimization approach implemented by \textit{Dual-CES}, which allows to produce focused summarizes with better saliency.

\subsubsection{Comparison with attentive baselines}
The pseudo-feedback distillation approach employed between the two steps of \textit{Dual-CES} has some resemblance to attention models that are used by state-of-the-art deep learning summarization methods~\cite{Cao2016AttSumJL,li2017cascaded,Ren:2017LCS}. First we note that, \textit{Dual-CES} significantly improves over these attentive baselines on ROUGE-1. On ROUGE-2, \textit{Dual-CES} is significantly better than \textit{C-Attention} and \textit{AttSum}, while it provides (more or less) similar quality to \textit{CRSum+SF}.

Closer analysis of the various attention strategies that are employed within these baselines, reveals that, while \textit{AttSum} only attends on a sentence representation level, \textit{C-Attention} and \textit{CRSum+SF} further attend on a word level. Such a more fine-granular attendance results in an improved saliency for the two latter. Yet, while \textit{C-Attention} first attends on sentences then on words, \textit{CRSum+SF} performs its attentions reversely. Using \textit{Dual-CES} as a reference method for comparison, apparently, \textit{CRSum+SF} attendance on salient words first and then on salient sentences based on such words seems as the better strategy.

In a sense, similar to \textit{CRSum+SF}, \textit{Dual-CES} also first ``attends" on salient words which are distilled from the pseudo-feedback reference summary. \textit{Dual-CES} then utilizes such salient words for better selection of salient sentences within its second step of focused summary production. Yet, compared to \textit{CRSum+SF} and similar to \textit{C-Attention}, \textit{Dual-CES}'s saliency ``attention" process is unsupervised. Moreover, \textit{Dual-CES} further ``attends" on salient sentence positions, which result in better tuning of the position-bias $b$ hyperparameter.

\begin{table}[tbh]
\small
\centering
\begin{tabular}{@{}llll@{}}
\toprule
$\bar{L}$               & R-1   & R-2   & R-SU4 \\ \midrule
500             & 45.52 & 12.32 & 17.69 \\
750             & 45.84 & 12.46 & 17.85 \\
1000            & 45.88 & 12.48 & 17.84 \\
1250            & 45.91 & 12.50 & 17.86 \\
1500            & \textbf{46.02} & \textbf{12.53} & \textbf{17.91} \\
1750            & 45.99 & 12.46 & 17.87 \\
2000            & 45.97 & 12.44 & 17.83 \\\hline
Adaptive-length ($L_t$) & 46.01 & 12.47 & 17.87 \\ \midrule

\end{tabular}
\caption{Sensitivity of \textit{Dual-CES} to the value of hyperparamter $\bar{L}$ (DUC 2007 benchmark)}\label{tab:sensitivity}
\end{table}

\begin{figure}[tbh]
  \centering
  \includegraphics[width=3.0in]{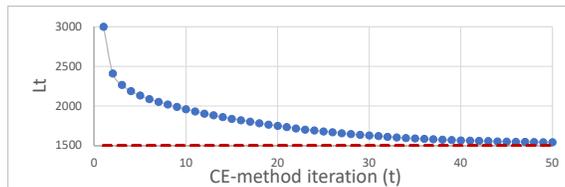}
  \caption{Illustration of the adaptive-length $L_{t}$ learning by \textit{Dual-CES-A} (DUC 2007 benchmark)}\label{fig:adaptive}
\end{figure}

\subsubsection{Hyperparamter $\bar{L}$ sensitivity analysis}

Table~\ref{tab:sensitivity} reports the sensitivity of \textit{Dual-CES} (measured by ROUGE-X Recall) to the value of hyperparameter $\bar{L}$, using the DUC 2007 benchmark. To this end, we ran \textit{Dual-CES} with an increasing $\bar{L}$ value. For further comparison, we also report in Table~\ref{tab:sensitivity} the results of its adaptive-length version \textit{Dual-CES-A}. \textit{Dual-CES-A} is still initialized with $L_{t=0}=3000$ and adaptively adjusts this hyperparameter. Figure~\ref{fig:adaptive} illustrates the (average) learning curve of its adaptive-length parameter $L_{t}$.

Overall, \textit{Dual-CES}'s summarization quality remains quite stable, exhibiting low sensitivity to $\bar{L}$. Similar stability was further observed for the two other DUC benchmarks. In addition, Figure~\ref{fig:adaptive} depicts an interesting empirical outcome: \textit{Dual-CES-A} converges (more or less) to the best hyperparameter $\bar{L}$ value (i.e., $\bar{L}=1500$ in Table~\ref{tab:sensitivity}). \textit{Dual-CES-A}, therefore, serves as a robust alternative for flexibly estimating such hyperparameter value during runtime. \textit{Dual-CES-A} can provide similar quality and may outperform \textit{Dual-CES}.

\section{Conclusions and Future work}\label{sec:conclusions}
We proposed \textit{Dual-CES}, an unsupervised, query-focused, extractive multi-document summarizer. 
\textit{Dual-CES} was shown to better handle the tradeoff between saliency and focus, providing the best summarization quality compared to other alternative state-of-the-art unsupervised summarizers. Moreover, in many cases, \textit{Dual-CES} even outperforms state-of-the-art supervised summarizers.
As a future work, we would like to learn to distill from additional pseudo-feedback sources. 

\balance

\bibliographystyle{plain}

\end{document}